# Deep Deconvolutional Networks for Scene Parsing

Rahul Mohan
Stanford University
Stanford, CA 94305
`rahulm@ai.stanford.edu`

## Abstract

*Scene parsing is an important and challenging problem in computer vision. It requires labeling each pixel in an image with the category it belongs to. Traditionally, it has been approached with hand-engineered features from color information in images. Recently convolutional neural networks (CNNs), which automatically learn hierarchies of features, have achieved record performance on the task. These approaches typically include a post-processing technique, such as superpixels, to produce the final labeling. In this paper, we propose a novel network architecture that combines deep deconvolutional neural networks with CNNs. Our experiments show that deconvolutional neural networks are capable of learning higher order image structure beyond edge primitives in comparison to CNNs. The new network architecture is employed for multi-patch training, introduced as part of this work. Multi-patch training makes it possible to effectively learn spatial priors from scenes. The proposed approach yields state-of-the-art performance on four scene parsing datasets, namely Stanford Background, SIFT Flow, CamVid, and KITTI. In addition, our system has the added advantage of having a training system that can be completely automated end-to-end without requiring any post-processing.*

## 1. Introduction

Scene parsing is one of the fundamental problems of computer vision. Scene parsing aims at segmenting images and detecting various object categories within them. Concretely, a scene parser classifies each pixel of an image into one of several predefined object classes. Like traditional computer vision systems such as detectors, a scene parsing model should ideally be robust to changes in illumination and viewpoint, and have an understanding of the spatial dependencies of the object classes in the images [36,37].

The parsing problem has been mainly addressed with Conditional Random Fields (CRFs) [8,9,35]. CRFs linearly combine input features (describing patches surrounding each pixel) along with contextual features (describing spatial interactions between labels) [8,9]. While these models have worked well on the scene parsing task [8], they require hand-engineered features. CRFs by themselves are also not able to capture large input contexts which is essential for detecting larger object classes such as road.

Recent work in this area has attempted to combine CNNs with CRFs to achieve state of the art results on many datasets [7]. The CNNs of LeCun et. al [24] are powerful feature detectors in which input data is subject to multiple layers of convolutions, non-linearities, and pooling (sub-sampling). One potential drawback of these networks is the use of spatial pooling in which mid-level cues such as edge intersections, parralelism, and symmetry are lost [5]. Such cues are very important in tasks such as scene parsing, especially when dealing with complex object classes, and therefore motivate the use of deconvolutional networks. Deconvolutional networks [5] are top-down models that learn features that capture such mid-level cues in image data. The model is completely unsupervised and seeks to reconstruct the input data using convolutional feature maps (instead of the input itself) and a set of learned filters, along with a sparsity constraint. Some of the interesting concepts that are able to be learned by deconvolutional networks are edge junctions, parallel lines, curves and basic geometric elements, such as rectangles [5].

The contributions of this paper are as follows: (1) The system introduces the use of deconvolutional networks in combination with traditional CNNs for feature learning and scene parsing. (2) It illustrates how llustrates how deconvolutional networks are able to learn more robust and inisghtful representations of the data when compared to regular CNNs. (3) A multi-patch training technique is introduced in order to learn an effective spatial prior. Additionally, the proposed system is automated and trained on raw pixels rather than using superpixels [38,39]. Section 2 details the methods used in our proposed system including training deconvolutional networks. Section 3 presents our experiments, results. Finally, our analysis of the network is given in Section 4.

## 2. Methods

### 2.1. Training Deconvolutional Layers

Deconvolutional Networks provide a conceptually simple unsupervised framework for learning sparse, overcomplete feature hierarchies [5,6,34]. Applying this framework to natural images produces a highly diverse set of filters that capture high-order image structure beyond edge primitives.

As with other deep learning models [27,28], deconvolutional networks look to learn hierarchies of features from data. Deep Belief Networks (DBNs) [8,32,33] and hierarchies of sparse auto-encoders [28, 29, 30], like deconvolutional networks, learn features in a greedy-layer wise unsupervised fashion.

In these approaches, each layer consists of an encoder and decoder [5,27,28,29,30]. The encoder maps the input space to the feature space while the decoder maps the latent features back to the input space, attempting to give a good reconstruction of the input. The input however is the output of the previous layer and not the original data.

Trying to extract a latent representation from the input without the use of an encoder is a difficult problem since it requires inference. There will be many parts in the latent feature space which will compete to try to explain various parts of the input. [5]

Deconvolutional networks do not use an encoder and rather directly solve an inference problem by using efficient optimization techniques described in [5]. The reason for this is that by computing the features exactly, better representations can possibly be learned. In addition, deconvolution networks attempt to reconstruct the input image rather than use the greedy-layer wise scheme in DBNs and stacked autoencoders.

We adopted the algorithm for training adaptive deconvolutional networks, presented by [5].

The goal of a deconvolutional layer $l$ is to find a feature map $z_l$ that minimizes a cost function $C_l(y)$ and best estimate a set of filters $f$. The cost function consists of the following: (i) a likelihood term that keeps the reconstruction of the input $\hat{y}$ close to the original input image $y$; (ii) a regularization term that penalizes the l1 norm of the 2D feature maps $z_l^k$ on which the reconstruction $\hat{y}_l$ depends. The relative weighting of the two terms is controlled by :

$$C_l(y) = \frac{\lambda_l}{2}\|\hat{y}_l - y\| + \sum_{k=1}^{K_l} |z_{k,l}|$$

First, the feature maps $z_l$ are convolved with a set of filters $f$. Specifically, the feature maps $z_l$ are latent variables specific to each image, which the filters $f$ are the parameters of the model common to all images. Next, $z_l$ is pooled using 3D max pooling and the pooled locations are saved in locations called switches (which are therefore binary matrices). The pooling operation can be defined as:

$$[p_l, s_l] = P(z_l)$$

where $p_l$ contains the pooled feature maps, $s_l$ contains the switches. Another type of pooling operation is done If $s$ is fixed, which take $s$ as an input and determine which elements of $z$ are copied into $p$. This can be defined as $p = P_s z$, where $P_s$ is a binary selection matrix set by the switches. Immediately after this, an *unpooling* operation is performed, which places the pooled values in the appropriate places in the original feature map and sets the rest of the elements to 0. The unpooled feature maps can be defined as $U_s$. A reconstruction operator $R_l$ is then defined that simply convolves ($F$) and unpools ($U_s$) them down to the input:

$$\hat{y}_l = F_1 U_{s,1}....F_l z_l = R_l z_l$$

We also define a projection operator $R_l^T$ that takes a signal as input and projects it back up to the feature maps of layer $l$, given previously determined switches $s_l...s_{l-1}$:

$$R_l^T = F_1 P_{s,1}....F_l P_s, l-1$$

After these operators and variables are defined, inference can be performed: a gradient step to update the feature maps $z_l$, a shrinkage steps that clamps small values in $z_l$ to 0 to increase sparsity, and updating of the switches $s_l$. The above 3 processes make up a single ISTA iteration.

Next, to estimate the filters $f$ in the model, which contain the parameters for the entire model, we take the derivatives of the cost function with respect to $f_l$ and set it to zero, and obtain the following linear system in $f_l$:

$$\sum_{i=1}^{N}(z_l^{i^T} P_{s_{l-1}}^i R_{l-1}^{i^T})\hat{y}_l^i = \sum_{i=1}^{N}(z_l^{i^T} P_{s_{l-1}}^i R_{l-1}^{i^T})y^i$$

where $\hat{y}_l^i$ is the reconstruction of the input using the current value of $f_l$. We solve this linear system using linear conjugate gradients (CG) to find the optimal set of filters. The full training algorithm is described in [5,6].

### 2.2. Network Architecture

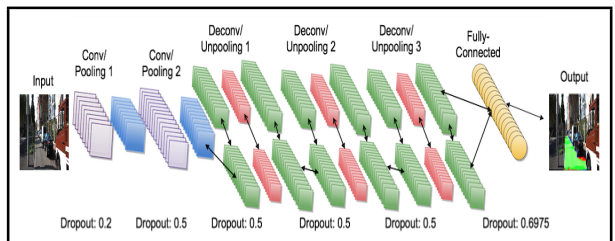

Figure 1. The architecture of our 7-layered deep network.

Our learning architecture combines feedforward layers with feedback layers. The training network consists of seven layers. The seven layers are made up of two convolutional layers in the beginning, followed by three deconvolutional layers, with a fully connected layer and softmax/sigmoid classifier at the top. Figure 1 illustrates the network architecture. We frame the scene parsing problem in the following way: there are $n$ input units and $m$ output units. Therefore, the softmax/sigmoid classifier works in an elementwise fashion where each output unit has a separate classifier. The convolutional layers and fully connected layers all used rectified linear as the nonlinearity function. Furthermore, we use dropout training [25] on each layer. Each layer in the network has a dropout rate of 0.5, except for the input layer which has a rate of 0.2 and the fully-connected layer which has a rate of 0.6975. We lowered the dropout rate on the input to avoid losing any information and increased the rate at the fully-connected layer since it is more susceptible to overfitting.

We apply convolution and max pooling on each of the convolutional stages. This is followed by a regular deconvolutional network on top, consisting of deconvolution and unpooling. We perform the learning in a sequential fashion. First, the two convolutional layers are trained to optimize the pixel-wise cross entropy through Stochastic Gradient Descent (SGD). Next, we switch the learning algorithm to use the Iterative Shrinkage Thresholding Algorithm (ISTA) [5,6] to train each of the deconvolutional layers. Finally, the fully-connected layer and the final classification output use SGD for updating their parameters.

The successive convolutions and max pooling operations in the inital 2 stages of the network enable the learning of useful initial hypothesis maps, thanks to added non-linearities and a larger spatial context. This part of the model itself is deep and produces complex representations of the input. The deconvolutional network on top provides feedback to these learned hypothesis maps and further improves these representations. This provides a very theoretically interesting framework combining deep supervised and unsupervised learning.

Furthermore, the deconvolutional layers could also be viewed as a form of fine-tuning. Fine-tuning is a supervised process to refine the weights of unsupervised models such as RBMs and autoencoders. The deconvolutional network in this architecture acts a type of *unsupervised* fine-tuning technique to refine the weights learned from the supervised 2-layered CNN in the bottom. Interestingly, this unsupervised fine-tuning technique is similar to the unsupervised pre-training technique [26] used with deep network; an experiment to demonstrate is presented in Section 4.

The small number of training examples and class imbalance for the datasets we tested on presented a significant risk of overfitting. This risk is increased by the large number of parameters used by our model. This problem was addressed in [1,3,7] by using class balancing techniques and including distortions of the input data. We followed the approach used in [1] and resampled the training data using balanced frequences such that an equal amount of each class is shown to the network. Training with balanced frequencies allows better discrimination of small objects and is more proper from a recognition point of view, although it may cause the pixel wise accuracy to reduce [7]. Finally, local contrast normalization was performed on each of the input datasets.

## 2.3. Multi-Patch Training

In scene parsing datasets, both the input image and output label are multi-dimensional vectors. Specifically, the input image is a three-dimensional vector with x and y coordinates of the pixels in the first two dimensions and the color channels in the third dimension. The output label is a two-dimensional vector with labels for each of the pixels. In order to obtain a good spatial prior to help learn even better features, we split the *output* images into multiple patches. During learning, we loop over each patch and train on the entire *input* image only to predict the pixels in that specific patch. The multi-patch training can also be thought of as having multiple networks which specialize at predicting a different part of the output. This is an effective way of breaking down the problem which is especially challenging given that we did not use superpixels to subsample the data.

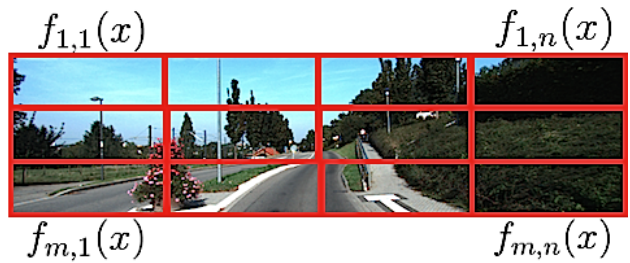

Figure 2. Example of multi-patch algorithm with $m$ rows and $n$ columns. Overall, there are $m \times n$ networks which each take as input $x$.

## 3. Experiments and Results

### 3.1. Datasets and Setup

We report our results on the following four datasets: Stanford Background, SIFT Flow, CamVid, and KITTI. The Stanford Background Dataset [11] contains 715 images of outdoor scenes and consists of 8 classes. The images were chosen from the publicly available datasets LabelMe, MSRC, PASCAL VOC and Geometric Context. We use the evaluation procedure described in [20,15] - five fold cross validation. Our only preprocessing step on all of the datasets was to normalize them to have zero mean and unit variance. The SIFT Flow dataset [12] is composed of 2,688 images, that have been thoroughly labeled by LabelMe users. Synonym correction was used by the authors to achieve 33 semantic labels. We use the same training and test split as [12]. The images are all 256 x 256 pixels. The Cambridge-driving Labeled Video Database (CamVid) dataset [13] contains ten minutes of video footage and corresponding semantically labeled groundtruth images at intervals. There are 11 semantic classes and 701 segmentation images. The video footage was recorded from driving around various urban settings and is particulary challenging given various lighting settings (dusk, dawn). The KITTI Vision Dataset [14], from the Honda Research Institute Europe GmbH, was created for drivable road detection. The images were extracted from video footage of driving around urban roads. The images were all manually annotated and contains binary labels of whether each pixel is drivable road or not. The website hosting the dataset does not provide the test labels and instead requires you to upload your test predictions to their server, providing you with your results after you have done so. The networks were trained on a computing cluster using multiple NVIDIA GPUs. Key hyperparameters (number of patches, patch size, number of deconvolutional filters, etc.) of the layers were tuned through cross validation. The settings and parameters for the experiments are all summarized in Tables 1, 2, and 3.

Table 1. Parameters and settings for experimental datasets.

| Dataset | Num. Train | Num. Test | Input Im. Size |
|---------|-----------|-----------|----------------|
| Stanford | 572 | 143 | 320 x 240 |
| SIFT Flow | 2,488 | 200 | 256 x 256 |
| CamVid | 468 | 233 | 320 x 240 |
| KITTI | 289 | 290 | 500 x 500 |

Table 2. Parameters and settings for experimental datasets.

| Dataset | Label Im. Size | Train Time | Patch Num. |
|---------|---------------|------------|------------|
| Stanford | 320 x 240 | 4 hrs 24 min | 16 |
| SIFT Flow | 256 x 256 | 6 hrs 8 min | 16 |
| CamVid | 320 x 240 | 3 hrs 52 min | 16 |
| KITTI | 375 x 1242 | 1 hr 43 min | 18 |

Table 3. Parameters and settings for experimental datasets. 'Parameters' and 'Ouput Units' are per-patch.

| Dataset | Patch Size | Parameters | Output Units |
|---------|-----------|------------|--------------|
| Stanford | 60 x 80 | 45 mil. | 4800 |
| SIFT Flow | 64 x 64 | 40 mil. | 4096 |
| CamVid | 60 x 80 | 45 mil. | 4800 |
| KITTI | 125 x 207 | 250 mil. | 25875 |

## 3.2. Overall Results

In this section, we present the results of our method on the four datasets mentioned in Section 3.1, in Tables 2 and 3. These results are from using the methods and architecture described in Section 2 and are therefore a good estimate of its generalizablity.

Table 4. Performance on the Stanford Background, SIFT Flow, and CamVid Datasets. For each of the datasets, we report per-pixel and average per-class accuracy, as in [7]. Our approaches and the best performing methods are bolded.

| Dataset | Approach | Pixel Acc. | Class Acc. |
|---------|----------|-----------|-----------|
| Stanford Background | Munoz et al. 2010 [15] | 76.9% | 66.2% |
| | Socher et al. 2011 [2] | 78.1% | — |
| | Pinheiro et a. 2014 [16] | 80.2% | 69.9% |
| | Farabet et al. 2013 [7] | 81.4% | 76.00% |
| | **Multi-Patch DeconvNet-8 Patches** | 83.92% | 77.86% |
| | **Multi-Patch DeconvNet-16 Patches** | **84.2%** | **78.37%** |
| SIFT Flow | Tighe and Lazebnik 2010 [3] | 77.0% | 30.1% |
| | Pinheiro et a. 2014 [16] | 77.7% | 30.0% |
| | Farabet et al. 2013 [7] | 78.5% | 29.6% |
| | Tighe and Lazebnik 2013 [17] | 78.6% | 39.2 |
| | **Multi-Patch DeconvNet-8 Patches** | 80.98% | 39.56% |
| | **Multi-Patch DeconvNet-16 Patches** | **81.67%** | **41.05%** |
| CamVid | Sturgess et al. [18] | 83.8% | 59.2 |
| | Floros et al. [19] | 83.2% | 59.6% |
| | Ladicky et al. [20] | 83.8% | 62.5% |
| | Tighe and Lazebnik [17] | 83.9% | 62.5 |
| | **Multi-Patch DeconvNet-8 Patches** | 84.53% | 63.83% |
| | **Multi-Patch DeconvNet-16 Patches** | **84.82%** | **64.1%** |

Table 5. Performance of our system on the KITTI Dataset. We report Maximum F1-measure (MaxF), Average Precision (AP), Precision (PRE), Recall (REC), False Positive Rate (FPR), False Negative Rate (FNR).

| Approach | MaxF | AP | PRE | REC | FPR | FNR |
|----------|------|-----|-----|-----|-----|-----|
| ProbBoost [21] | 87.21 % | 77.79 % | 86.96 % | 87.47 % | 7.55 % | 12.53 % |
| SPRAY [22] | 86.33 % | **90.91 %** | 86.78 % | 85.89 % | 7.53 % | 14.11 |
| RES3D-Velo [23] | 85.49 % | 79.03 % | 79.93 % | **91.88 %** | 13.28 % | **8.12** |
| GRES3D-V | 83.97 % | 78.46 % | 79.91 % | 88.46 % | 12.81 % | 11.54 % |
| CB | 83.83 % | 88.44 % | 83.01 % | 84.67 % | 9.98 % | 15.33 % |
| **Multi-Patch DeconvNet-8 Patches** | 92.08 % | 89.14 % | 93.53 % | 90.19 % | 3.74 % | 10.24 % |
| **Multi-Patch DeconvNet-16 Patches** | **92.51 %** | 89.36 % | **94.64 %** | 90.46 % | **2.95 %** | 9.54 % |

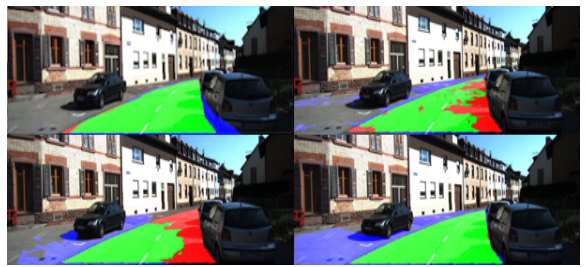

Figure 3. Drivable road predictions on the KITTI Dataset compared with 3 next best approaches. **Top Left: Our Approach**, Top Right: SPRAY [21], Bottom Left: ProbBoost [22], Bottom Right: RES3D-Velo [23]. In the images, green signifies true positives. Blue denotes false positives while red denotes false negatives. As shown, our predictions are more consistent and clean.

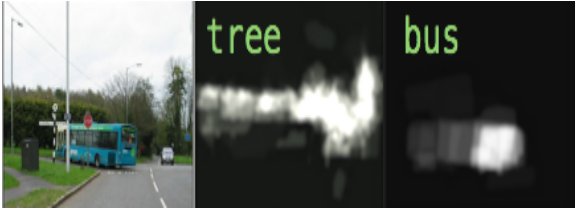

Figure 4. The heatmap for the tree and bus class on an image from the SIFT Flow dataset. Specifically, the probability of those particular classes at each pixel in the image is shown.

# 4. Analysis

Our model achieved state-of-the-art performance in both per-pixel and average per-class accuracy, on the Stanford Background, SIFT Flow, and CamVid datasets. Furthermore, the model achieved the best-recorded result on the KITTI dataset to date, as it obtained a state-of-the-art maximum F-score by over 5%. Overall, the model's consistency and effectivness over a variety of datasets led us to investigate why it works so well. In this section, we present various different experiments and observations and try to pinpoint the biggest factors leading to the model's performance.

### 4.0.1    Performance and Generalization of our Network

We conducted several experiments to test the effect of the deconvolutional stage in our architecture. Our first experiment was to remove each deconvolutional layer one-by-one and re-measure the generalization accuracy on each of the datasets. This allows us to measure the importance of depth along with amount of contribution from the deconvolutional layers. In this experiment, the network where all of the deconvolutional layers are removed results in a two-layered CNN. In order to avoid bias while comparing the plain CNNs to the deconvolutional networks, we did the following: (1) Made sure the number of total parameters of the CNN was equal to that of the corresponding deconvolutional network. (2) Used the same settings for the CNNs (i.e rectified linear hidden layers, hyperparameter optimization through cross validation, same number of patches and dropout amount).

As shown in Figure 5, a deep hierarchy of deconvolutional layers greatly increases the performance of the network. Specifically, the average difference in per-pixel accuracy of the network with all 3 deconvolutional layers (Deconv-5) and the network with no deconvolutional layers (CNN-2) is **5.46%**, on all of the datasets besides KITTI. On KITTI, the difference between those two network's Max F-Scores is **7.3%**. In fact, if all of the deconvolutional layers are removed, the resulting accuracies and Max F-Score falls below the previous best for each of the datasets. Furthermore, each addtional deconvolution layer adds on average

**1.16%** to the per-pixel accuracy, on all of the datasets besides KITTI. On KITTI, each addtional deconvolution layer adds on average **1.86%** to the Max F-Score.

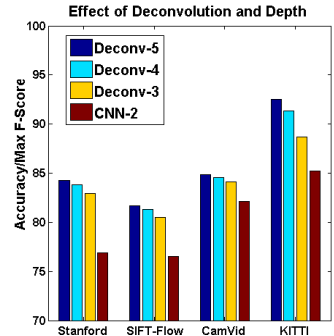

Figure 5. Results of our first experiment in which we removed deconvolutional layers one-by-one. 'Deconv-5' refers to the original network with 3 deconvolutional layers (the third one being the 5th layer of the network) and so forth. 'CNN-2' refers to all the deconvolutional layers being removed.

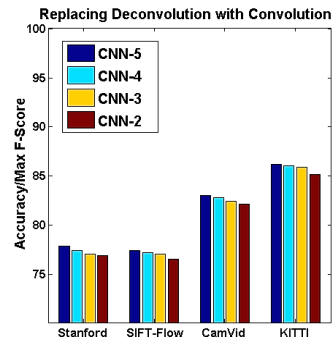

Figure 6. Results of our second experiment in which we replaced the deconvolutional layers with plain convolutional layers. 'CNN-5' refers to the original network with 5 convolutional layers (instead of 2 convolutional layers and 3 deconvolutional layers) and so forth. 'CNN-2' is the same as in Figure 4.

In order to verify that the deconvolutional layers were truly effective, we conducted a second experiment where *replaced* the deconvolution layers with regular convolutional layers. Therefore, instead of a network with two convolutional layers and three deconvolutional layers, we use a network with five convolutional layers. We also experiment with shallower pure convolutional architectures to directly compare with the experiments in Section 3.2.1. Figure 6 shows the results for four different convolutional networks we tested. The two-layered network (CNN-2) is equivalent to the network shown in Figure 5 where all of the deconvolutional layers were removed.

The CNN-5 shown in Figure 6 is not able to achieve state-of-the-art accuracy on any of the datasets. Further-more, the difference between the CNN-5 and CNN-2 in per-pixel accuracy is on average **0.78%**. Furthermore, the difference between CNN-5 and CNN-2 on KITTI is only **0.52%** in Max F-Score. Additionally, each additional convolutional layer adds on average only about **0.41%** to the per-pixel accuracy, and **0.22%** to the Max F-Score on KITTI. All of these numbers are far less than those obtained with Deconv-5, which are mentioned above.

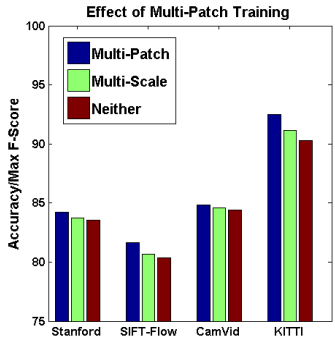

Figure 7. Results of our experiment with multi-patch training which is compared with multi-scale learning and no special training method at all. We extract 4 scales for the multi-scale learning using a Laplacian pyramid [7].

The results of these experiments demonstrates the effectiveness of deconvolutional networks and shows that they are capable of learning more informative features than CNN's themselves, when stacked with CNNs in a deep hierarchy. The depth of the deconvolutional stage in our network is also far more important when compared with regular CNNs. To investigate this further, we visualized the filters of each layer from the deconvolutional network. Figure 8 shows the visualization of the filters from the last convolutional layer and all 3 deconvolutional layers in the network. As expected, the filters are more complicated and represent higher level features as the depth increases. The filters in the deconvolutional layers also go beyond edge primitives.

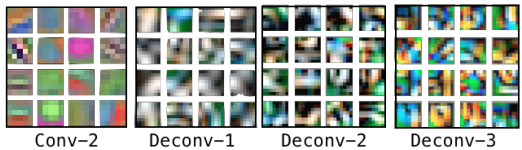

Figure 8. Filters from the last convolutional layer and all of the deconvolutional layers. As shown, the filters get significantly more complicated by layer in the deconvolutional layers.

We were also able to prove that our network architecture is more stable than CNNs and significantly reduces the problem of getting stuck in poor apparent local minima. Specifically, we show how increasing the depth of CNNs increases the probability of finding poor local minima when starting from random initialization seeds, while our network generally is robust and avoids such local minima even when it's architecture is very deep. We use the SIFT Flow dataset for this experiment as it is the most challenging out of the four used. We run Deconv-5 and CNN-5 along with Deconv-3 and CNN-3 500 different times (with 500 different random initialization seeds) each and record the pixel-wise accuracy during each run.

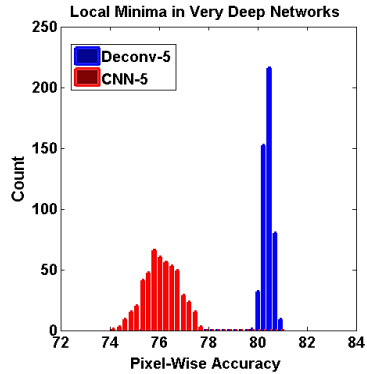

Figure 9. 500 random runs with Deconv-5 and CNN-5.

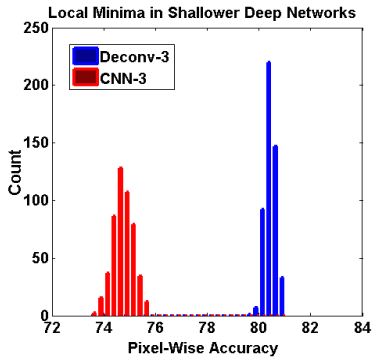

Figure 10. 500 random runs with Deconv-3 and CNN-3. While the variance of CNN-5 is much higher than CNN-3, Deconv-5 and Deconv-3 are very consistent.

As shown in Figure 9 and 10, our network is robust with respect to the random initialization seed as well as depth. The variance and the number of bad outliers is higher with CNN-5 when compared to CNN-3, representing a common problem when training very deep architectures on small datasets. On the other hand, Deconv-5 and Deconv-3 both have similar variances that are much lower than those obtained by the CNNs suggesting the network is arriving at better optima on a much more consistent basis. Interestingly, Erhan. et al [26] pointed out that unsupervised pre-training is also robust with respect to the random initialization seed. Since deconvolutional networks can be thought of as a form of unsupervised fine-tuning, there are similarities between the two methods.

Our network is a step forward in optimizing large deep networks better, especially on the small datasets such as those used in this paper.

### 4.0.2 Effect of Multi-Patch Training

Next, we assesed the importance of our multi-patch training technique. To do so, we compared it *multi-scale* learning and no special training procedure at all. Multi-scale learning, in the context of raw scene parsing, involves taking various scales or crops centered around pixels and learning features seperately on them [1]. The features are all then concatenated and fed to a classifer (e.g. softmax).Multi-scale learning has achieved great success on scene parsing [1]. Figure 7 summarizes the results for each of the datasets. Overall, our multi-patch training technique outperformed multi-scale learning and no special training methods, on every dataset. Specifically, on average the multi-patch training adds on **0.67%** more than multi-scale training to the per-pixel accuracy, on each dataset besides KITTI. On KITTI, the multi-patch training adds on **1.62%** more than multi-scale training to the Max F-Score. Furthermore, when compared with no special training method, multi-patch training adds on average **1.02%** more to per-pixel accuracy on the first three datasets and **3.28%** to the Max F-Score on the KITTI dataset. Overall, while the multi-patch training did help slightly, we can conclude that the deconvolutional layers had the largest impact in the network.

## 5. Conclusion

This paper presents a new network architecture for scene parsing that combines deep deconvolutional and convolutional neural networks. It also introduces a multi-patch training technique that is capable of learning an effective spatial prior. We demonstrate how our network combined with multi-patch training produces state of the art results on multiple scene parsing datasets by a significant margin. The model is also completely automated as it takes in raw pixels and does not use any pixel-subsampling techniques such as superpixels. Our work shows the promise of using this network architecture to improve performance on other computer vision tasks and learn better hierarchies of features.

## 6. Acknowledgements